\documentclass[conference]{IEEEtran}
\usepackage[suppress]{color-edits}
\usepackage{amsmath,amssymb,amsfonts, amsthm}
\usepackage{algorithmic}
\usepackage{graphicx}
\usepackage{textcomp}
\usepackage{xcolor}
\usepackage[numbers]{natbib}
\usepackage[hyphens]{url}
\usepackage{hyperref}
\definecolor{deeppink}{rgb}{1.0, 0.08, 0.58}
\hypersetup{
  colorlinks = true, 
  urlcolor   = deeppink, 
  linkcolor  = blue, 
  citecolor  = red 
}
\usepackage{lipsum}
\newcommand\blfootnote[1]{%
  \begingroup
  \renewcommand\thefootnote{}\footnote{#1}%
  \addtocounter{footnote}{-1}%
  \endgroup
}

\newtheorem{definition}{Definition}

\addauthor[Hoda]{hh}{red}
\addauthor[Anna]{ak}{purple}
\addauthor[Ken]{kh}{violet}
\addauthor[Amanda]{ac}{brown}
\addauthor[Amanda]{am}{brown}
\addauthor[Haiyi]{hz}{blue}

\def\BibTeX{{\rm B\kern-.05em{\sc i\kern-.025em b}\kern-.08em
    T\kern-.1667em\lower.7ex\hbox{E}\kern-.125emX}}

\begin{document}
\newcommand{\xhdr}[1]{\vspace{1mm} \noindent{\bf #1}.\:}

\title{A Validity Perspective on Evaluating the Justified Use of Data-driven Decision-making Algorithms}

\author{\IEEEauthorblockN{Amanda Coston}
\IEEEauthorblockA{\textit{Heinz College \& Machine Learning Dept.} \\
\textit{Carnegie Mellon University}\\
Pittsburgh, USA\\
\href{mailto:acoston@cs.cmu.edu}{\texttt{acoston@cs.cmu.edu}}}
\vspace{1.2em}
\IEEEauthorblockN{Ken Holstein}
\IEEEauthorblockA{\textit{Human-Computer Interaction Institute} \\
\textit{Carnegie Mellon University}\\
Pittsburgh, USA\\
\href{mailto:kjholste@cs.cmu.edu}{\texttt{kjholste@cs.cmu.edu} }} 
\and
\IEEEauthorblockN{Anna Kawakami}
\IEEEauthorblockA{\textit{Human-Computer Interaction Institute} \\
\textit{Carnegie Mellon University}\\
Pittsburgh, USA\\
\href{mailto:akawakam@cs.cmu.edu}{\texttt{akawakam@andrew.cmu.edu}} }
\and
\IEEEauthorblockN{Haiyi Zhu}
\IEEEauthorblockA{\textit{Human-Computer Interaction Institute} \\
\textit{Carnegie Mellon University}\\
Pittsburgh, USA\\
\href{mailto:haiyiz@cs.cmu.edu}{\texttt{haiyiz@cs.cmu.edu}} }
\vspace{1.2em}
\IEEEauthorblockN{Hoda Heidari}
\IEEEauthorblockA{\textit{Machine Learning Dept.} \\
\textit{Carnegie Mellon University}\\
Pittsburgh, USA\\
\href{mailto:hheidari@cmu.edu}{\texttt{hheidari@cs.cmu.edu}} }
 }

\maketitle

\begin{abstract}
Recent research increasingly brings to question the appropriateness of using predictive tools in complex, real-world tasks. While a growing body of work has explored ways to improve value alignment in these tools, comparatively less work has centered concerns around the fundamental justifiability of using these tools. This work seeks to center validity considerations in deliberations around whether and how to build data-driven algorithms in high-stakes domains. Toward this end, we translate key concepts from validity theory to predictive algorithms. We apply the lens of validity to re-examine common challenges in problem formulation and data issues that jeopardize the justifiability of using predictive algorithms and connect these challenges to the social science discourse around validity. Our interdisciplinary exposition clarifies how these concepts apply to algorithmic decision making contexts. 
We demonstrate how these validity considerations could distill into a series of high-level questions intended to promote and document reflections on the legitimacy of the predictive task and the suitability of the data.\blfootnote{This work was generously funded by the National Science Foundation Graduate Research Fellowship Program under Grant No. DGE1745016, support from PwC and from CMU Block Center for Technology and Society Award No. 53680.1.5007718. Any opinions, findings, and conclusions or recommendations expressed in this material are solely those of the authors.}
\end{abstract}


\section{Introduction}
Data-driven algorithmic decision-making, in theory, can afford improvements in efficiency and the benefits of evidence-based decision making. Yet in practice, data-driven decision systems, often taking the form of algorithmic risk assessments, have caused significant adverse consequences in high-stakes settings. Investigators have identified unintended and often biased behavior in algorithmic decision systems used in a variety of applications, from detecting unemployment and welfare fraud to determining pre-trial release decisions and child welfare screening decisions, as well as  in algorithms used to inform medical care and set insurance premiums \cite{eubanks2018automating, angwin2016machine, obermeyer2019dissecting, vyas2020hidden, gilman2020welfare, charette2018unemployment, angwin2016premium, fabris2021algorithmic}. 
These high-profile incidents have brought into focus key questions such as how we can anticipate these harms before deployment\hhedit{, and perhaps more fundamentally,} whether algorithms are \hhdelete{even }suitable \hhedit{in the first place} for \hhedit{such high-stakes} decision-making tasks. 

In this work, we examine how validity considerations can help guide decisions about whether to build and deploy algorithmic decision systems.
\hhdelete{As our protocol is intended to inform the decision to deploy \hhcomment{Aren't we also interested in the question of 'whether to build'?}, it is relevant to }
\hhedit{Our proposal can be }contextualize\hhedit{d} \hhdelete{our proposal }in the tradition of technology refusal. Activists have long argued for the value in refusing technology and opting not to build \cite{tierney2019dismantlings, baumer2011whennot}. These calls have taken on new urgency in the modern setting of algorithmic proliferation as many scholars and activists debate
when to repair or abolish the use of algorithms in socially consequential settings \cite{kluttzafog, roberts2019digitizing, benjamin2019race, selbst2019fairness, abebe2020roles, barocas2020not, minow2019technical}.  

To anticipate harms before deployment, researchers and practitioners have proposed a suite of tools and processes\hhdelete{ for responsible AI}. This work has \khedit{frequently} considered questions of value-alignment, such as how to promote fairness and establish transparency and accountability \cite{uk_data_ethics_framework, madaio2020co, raji2020closing, mitchell2019model, gebru2021datasheets}. More recently, there have been growing calls to assess the appropriateness of using predictive tools for complex, real-world tasks from a \textit{validity} perspective \cite{raji2022fallacy}. In many cases where algorithms prove unsuitable for real-world use, the problem originates in the initial problem formulation stages \cite{passi2019problem, barocas2016big}\khedit{, or in the process of operationalizing latent constructs of interest (e.g., worker well-being, risk of recidivism, or socioeconomic status) via more readily observable measures and indicators \cite{jacobs2021measurement, narayanan2019recognize, recht_2022}}. Without addressing these issues directly, it may be challenging or impossible to \hhdelete{re-align values}\hhedit{align the resulting model with human values} after the fact. In some cases, efforts to do \hhedit{so} may actually backfire because of unaddressed upstream issues.

Our work seeks to center validity considerations,  a crucial criterion for the \hhdelete{responsible}\hhedit{justified} use of algorithmic \khedit{tools in real-world} decision-making \khedit{\cite{jacobs2021measurement, narayanan2019recognize, recht_2022}}. In doing so, we situate our work at the intersection of \khedit{research that debates}\khdelete{the community that debates} algorithm refusal \akedit{versus} repair and \khedit{research that develops}\khdelete{the community that develops} artifacts for responsible AI/ML.  
Guided by the goal of delivering an accessible tool to promote deliberation and reflection around validity, we \amedit{propose a structure for a protocol designed to distill} common validity issues into a question-and-answer (Q\&A) format. 
\amdelete{\khedit{Longer term, we}\khdelete{We} \hhedit{envision}\hhdelete{propose} combining this question bank with existing tools designed for \khedit{AI/ML} fairness and ethics into a composite protocol that is styled after Institutional Review Board (IRB) oversight for human subjects research.}

The main contributions of this paper are as follows:
\begin{enumerate}
    \item We provide a working taxonomy of criteria for the justified use of algorithms in high-stakes settings. We utilize this taxonomy to illuminate two \amedit{important} principles for \hhedit{substantiating/refuting the use of ML for decision making}\hhdelete{responsible machine learning}: validity and reliability (Section~\ref{section: taxonomy}).  \amdelete{Our paper aims to address validity.}
    \amedit{ \item  We use this taxonomy to conduct an interdisciplinary literature review on validity, reliability, and value-alignment (Section~\ref{section: related}).
    \item We connect modern validity theory from the social sciences to  common challenges in problem formulation and data issues that jeopardize the validity of predictive algorithms in decision making (Section~\ref{section: challenges}).
    \item We demonstrate how this systematization can inform future work by sketching the structure for a protocol to promote deliberation on validity. \amdelete{, which we plan to further develop in collaboration with  \akedit{frontline decision-makers}, ML practitioners\hhdelete{ (e.g., in public sector decision-making contexts)}, impacted community members, and other relevant stakeholders (Section~\ref{section: deliberation}).}
    }
    \amdelete{
    \item We distill these issues into \hhedit{an initial set of} questions designed to promote deliberation on validity\khedit{, which we plan to further develop in collaboration with ML practitioners (e.g., in public sector decision-making contexts), community members who are impacted by ML systems, and other relevant stakeholders}. }
\end{enumerate}

Throughout the paper we will discuss validity in the context of several high-stakes settings where predictive algorithms are increasingly used to inform human decisions: pre-trial release in the criminal justice system and screening decisions in the child welfare system. In the criminal justice setting, judges must decide whether to release a defendant before trial based on the likelihood that, if released, the defendant will fail to appear for trial as well as the likelihood the defendant will be arrested for a new crime before trial \cite{kleinberg2018human}.
For the child welfare screening task, call workers must decide which reports of alleged child abuse or neglect should be screened in for investigation based on an assessment of the likelihood of immediate danger or long-term neglect if no further action is taken \cite{chouldechova2018case}.

\hhdelete{\section{Principles for responsible data science: a working taxonomy}}
\hhedit{\section{A Taxonomy of Criteria for Justified-Use of Data-driven Algorithms}\label{section: taxonomy}}

\hhcomment{We need to make the rationale behind the taxonomy crystal clear.}

\hhedit{To assess whether the use of data-driven algorithms is adequately justified in a given decision making context, one must account for a wide range of factors. To give structure to this vast array of considerations, we propose a high-level taxonomy---we posit that the justified use of algorithmic tools \khedit{requires \textit{at minimum}}\khdelete{least} account\khedit{ing} for validity, value-alignment, and reliability. In this section, we offer a precise definition for these terms. Section~\ref{section: related} offers an overview of existing literature on each of these topics. }

\hhedit{
\paragraph{The rationale for our taxonomy:} To evaluate whether the use of predictive tools is sufficiently justified in a high-stakes decision making domain, at a minimum, we need to answer the following sequence of questions:
\begin{itemize}
    \item Can we translate (parts of) the decision making task into a prediction problem where both a measure representing the construct we'd like to predict and predictive attributes are available in the observed data?  
    \item If the answer to the above question is affirmative, does the model we train align with stakeholders' values, such as impartiality and non-discrimination?
    \item Do we understand the longer-term consequences of deploying the model in decision making processes? For example, how might the deployment setting change over time and can the model be reliably utilized under this changing environment?
\end{itemize}
The above questions motivate our three high-level categories of considerations for justifying/refuting the use of data-driven algorithms in decision making: validity, value alignment, and reliability.
}

\hhedit{Before we elaborate on our taxonomy, two remarks are in order. First, we emphasize that a} formal, comprehensive taxonomy \hhedit{of considerations around justified-use of algorithms is} a formidable research question in itself, and \hhdelete{we emphasize that we describe a working taxonomy for the purposes of} the purpose of our taxonomy is limited to structuring our review of the available literature, tools and resources\hhdelete{ on responsible data science}. We make no claims \khedit{regarding the comprehensiveness of our taxonomy.}\khdelete{on our taxonomy being optimal or complete.} We refer the interested reader to treatises on the subject including \citet{fjeld2020principled, floridi2021unified, golbin2020responsible}. \hhedit{Additionally, we note that the three categories at the heart of our taxonomy are intimately connected, rather than mutually exclusive.\khdelete{not mutually exclusive, but rather are intimately connected.}} 

\xhdr{Validity}
\hhdelete{Our starting point for responsible data science requires establishing}\hhedit{Our first category of considerations, validity, aims to establish that} the system does what it purports to do.
This quality is much harder to satisfy than one might initially think. Consider for instance the task of predicting which criminal defendants are likely to reoffend. Predictive models are often trained using re-arrest outcomes \cite{fogliato2021validity}.
Whether a model predicting re-arrest actually predicts reoffense is subject to considerable debate, particularly given that a large body of work has established racial disparities in arrests even for crimes which have little differences in prevalence by race \cite{alexander2011new}. A model that appears accurate with respect to re-arrests may be quite inaccurate with respect to actual crime. 
\acdelete{When our system does something other than what we expect it to, this can potentially invalidate all further consideration of responsible practices.} More broadly, the notion of validity requires not only that
 the system has to predict what it purports to predict, but also must achieve acceptable accuracy both within and outside the training environment (in the real-world deployment). These validity criteria are adapted from validity considerations (e.g., \textit{construct validity}, \textit{internal validity}, and \textit{external validity}) that are widely adopted in social sciences, including psychology, psychometrics, and Human-Computer Interaction \cite{campbell1957factors,messick1995validity,gergle2014experimental}.


\begin{definition}[Validity]
A measure, test, or model is valid if it closely reflects or assesses the specific concept/construct that the designer intends to measure~\citep{drost2011validity}. 
\end{definition}

We say that a predictive algorithm is valid when it predicts the quantity that we think it does, and similarly we say that an audit or assessment is valid when it evaluates the quantity that we would like to audit or assess. Threats to validity can arise as early as the problem formulation stage where decisions about how to operationalize the problem can induce misalignment between what we intend to predict versus what the model actually predicts \citep{passi2019problem, jacobs2021measurement}. When validity does not hold, it is quite challenging to  assess value-alignment---our next category of considerations. In this sense, we claim that validity is a prerequisite for the more commonly discussed values such as fairness.

\hhedit{\xhdr{Value-alignment} Our second category of considerations focuses on the compliance of the system with stakeholders' values.}
\begin{definition}[Value-alignment]
Value-alignment requires that the goals and behavior of the system comply with \hhedit{collective}\hhdelete{social} values \khedit{of relevant stakeholders and communities \citep{sierra2021value}.} 
\end{definition}
\hhedit{Relevant stakeholders might include the communities that will impacted by the algorithmic system or the frontline workers who will work with the system.}
Commonly discussed values include fairness, \hhdelete{trust, human dignity, justice,} privacy, transparency, and accountability. 
Properties like simplicity and interpretability are often desired as a means to ensure these values \citep{rudin2020age}, and within this taxonomy, we include these properties under the broad umbrella of value-alignment. 

\xhdr{Reliability}
The final set of considerations that we will discuss concern \hhedit{reliability} over time and context. 
\begin{definition}[Reliability]
\hhedit{Reliability is the extent to which the output of a measurement/test/model is \emph{repeatable, consistent, and stable} --- when different persons utilize it, on different occasions, under different conditions, with \hhdelete{supposedly} alternative instruments \hhedit{that} measure the same thing \citep{drost2011validity}.}
\hhdelete{Reliability requires that the validity and value-alignment of the system are robust to changes over time and context.}
\end{definition}
%
%
Reliability concerns in part the dynamical nature of systems in the real world. A system that satisfies our \hhedit{previous two} criteria at a given snapshot in time may soon after experience a policy, population, or other notable change that may have profound effects on its validity and value-alignment. \hhdelete{We use the term \emph{reliability} to refer to the quality that a system's validity and value-alignment are robust to changes over time.}
Threats to reliability include changes in the population characteristics and/or risk profiles (i.e., distribution shift) or strategic behavior in response to the algorithmic model predictions.

We use this taxonomy to structure a literature review of related work in the following section.

\section{Literature review}\label{section: related}
In this section we conduct a structured literature review of prior work in validity, value-alignment, and reliability.


\begin{figure}
    \includegraphics[scale=0.50]{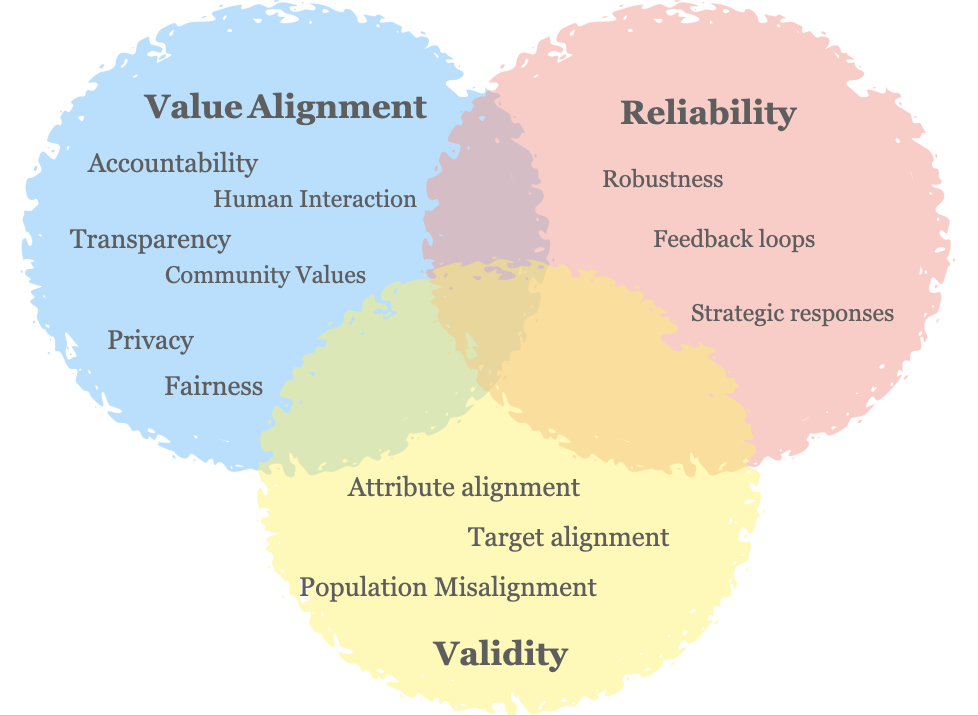}
    \caption{The justified use of algorithms in high-stakes decision making requires at minimum that we account for validity, reliability and value alignment. These concepts are overlapping and interconnected, encompassing many aspects of responsible machine learning.} 
     \label{fig:framework}
\end{figure}

\subsection{Validity}
We begin our literature review with validity. The machine learning literature has vibrant communities addressing validity-related considerations, such as selection bias and representation bias, but, to the best of our knowledge, there is no unifying validity framework around these issues. For this we turn to the theory of validity in the social sciences. In this section we review key concepts from social science research on validity, and in subsequent sections we translate these concepts to the setting of data-driven algorithms.  

 \hzedit{\textbf{Construct validity} is concerned with whether the measure captures what the researcher intended to measure. Modern validity theory often defines construct validity as the overarching concern of validity research: construct validity integrates considerations of content, criteria, and consequences into a unified construct framework \cite{messick1995validity, schotte1997construct}. \citet{messick1995validity} and \citet{gergle2014experimental} highlight distinguishable aspects of construct validity. Below we review the definition of different aspects of construct validity\khdelete{ and highlight the aspects that are most}\khedit{, highlighting aspects that are particularly} relevant in assessing the validity of data-driven decision-making algorithm. 
\begin{itemize}
    \item \textbf{\textit{Face validity}} means that the chosen measure ``appears to measure what it is supposed to measure" \cite{gergle2014experimental}. For example, imagine you propose to assess or predict the online satisfaction with a product on a e-commerce website by measuring the proportion of positive comments among all the purchase comments. You feel that the higher the proportion of the positive comments, the more satisfied the customers were, so “on its face” it is a valid measure or prediction target. Face validity is a very weak requirement and should be used analogously to rejecting the null in hypothesis testing: rejecting face validity allows us to conclude that the measure is not valid, but failure to reject face validity does not allow us to conclude it is valid.
    \khdelete{\khedit{\item \textbf{\textit{Content validity}} means that ............................}}
    \item \textbf{\textit{Convergent validity}} uses more than one measure for the same construct and then demonstrates a correlation between the two measures at the same point in time. One common way to examine convergent validity is to compare your measure with a gold-standard measure or benchmark. 
    However, \citet{gergle2014experimental} warned that convergent validity can suffer from the fact that the secondary variable for comparison may have similar limitations as the measure under investigation.
    \item \textbf{\textit{Discriminant validity}} tests whether measurements of two concepts that are supposed to be unrelated are, in fact, unrelated. Historically researchers have struggled to demonstrate discriminant validity for measures of social intelligence because these measures correlate highly with measures of mental alertness \citep{campbell1959convergent}.
    \item \textbf{\textit{Predictive validity}} is a validation approach where the measure is shown to accurately predict some other conceptually related variable later in time. For example, in the context of child welfare, \citet{vaithianathan2020hospital} demonstrated the predictive validity of Allegheny Family Screening Tool (AFST) by showing that the AFST's home removal risk score \textit{at the time of a maltreatment referral}, was also sensitive to identifying children with a heightened risk of an emergency department (ED) visit or hospitalization because of injury \textit{during the follow-up period}. Therefore, they argued ``the risk of placement into foster care as a reasonable proxy for child harm and therefore a credible outcome for training risk stratification models for use by CPS systems" \cite{vaithianathan2020hospital}.
\end{itemize}
}

\hzedit{\textbf{Internal validity} and \textbf{external validity} are important validity considerations in experimental research \cite{campbell1957factors, gergle2014experimental}. Internal validity is the degree to which the claims of a study hold true for the particular (often artificial) study setting, while external validity is
the degree to which the claims hold true for real-world contexts, with varying cultures, different population, different technological configurations, or varying times of the day \cite{gergle2014experimental}. \citet{gergle2014experimental} discussed three common ways to bolster external validity in study design: (1) choosing a study task that is a good match for the kinds of activities in the field, (2) choosing participants for the study that are as close as possible to those in the field, and (3) assessing the similarity of the behaviors between the laboratory study and the fieldwork.
}

Prior work on data-driven decision-making algorithms has probed various aspects of validity threats or concerns, \hzedit{often using the vocabulary of} ``measurement error", ``problem formulation", and ``biases". For example, \citet{passi2019problem} chronicle how the analysts' decisions during problem formulation impacts fairness of the downstream model. Relatedly \citet{jacobs2021measurement} demonstrate that how one operationalizes theoretical constructs into measurable quantities impacts fairness. \citet{suresh2021framework} also highlight measurement error in their characterization of seven types of  harm in machine learning and describe other biases in representation and evaluation that can threaten validity. Representation and evaluation biases can occur when the development sample and evaluation sample, respectively, do not accurately represent who is in the target population. 
To the best of our knowledge, there is no prior work that proposes tools or processes centered around validity issues. Our paper aims to fill this gap by drawing on the findings in these papers to structure a validity-centered artifact intended for real-world use.

\hhcomment{The following section is relatively long. Ideally our validity subsection should be the most substantial of the three. Can we interpret some of the material under value alignment as validity considerations?}
\accomment{That's a good point about the section being too long. Perhaps I can first see how to tighten the section (I originally wrote it with a lot of detail for our own internal purposes but that is probably too much for a submission). We make the case there aren't artifacts designed for validity, so it feels better to keep them in the value-alignment section if possible. }

\subsection{Value Alignment}
 The literature on value-alignment is vast, and we therefore focus on the works most related to our purpose of developing \emph{artifacts}, such as documents, checklists, and software toolkits, to promote \hhdelete{responsible}\hhedit{justifying the} use of algorithmic systems in decision-making. 
\amedit{Documentation artifacts designed to improve transparency and \akedit{inform} trust have been proposed for datasets, machine learning models, and AI products and services \cite{holland2020dataset, gebru2021datasheets, hutchinson2021towards, mitchell2019model, arnold2019factsheets}. These artifacts document typical use cases, product/development lineage, and other important specificatons in order to promote proper use as the models, data, and services are shared and re-used across a variety of contexts.}
Noticing that these documentation products largely represent the perspective of algorithm developers, a recent work developed a toolkit designed to engage community advocates and activists in this process \citep{krafft2021aekit}. \acdelete{The Algorithmic Equity Toolkit (AEKit) provides a flowchart to determine if a service is an automated decision system or surveillance system, a questionnaire to identify potential harms, and a worksheet to document stakeholder concerns and potential misuses of the system. The AEKit provides structure for a community member without inside knowledge of the data or algorithm to participate in the oversight process. Another mechanism for representing community stakeholder perspectives and concern is through research papers that summarize findings from workshops with community members. This research has solicited community feedback on algorithms used in industry \citep{woodruff2018qualitative} and in public policy settings \cite{brown2019toward, cheng2021soliciting}.}

An increasingly popular mechanism is checklists for fairness and ethics in machine learning.
 Checklists can provide a structured form for individual advocates to raise fairness or ethics concerns, but a compliance-oriented checklist may fail to capture the nuances of complex fairness and ethical challenges \cite{madaio2020co}. Recent work has advocated for checklists designed to promote conversations about ethical challenges \cite{madaioprompting}. However, checklist-style ``yes or no" questions may be ill-suited for promoting deliberation.  \amedit{
 Moreover, in centering around the question ``\emph{have we performed all the steps necessary before releasing the model?}''}, checklists adopt a ``deploy by default" framing that may encourage practitioners to err on the side of brushing concerns aside. To address these issues, we sketch a protocol to promote deliberation centered around the question \emph{``is an algorithmic model appropriate for use in this setting?''}.
 
\citet{raji2020closing} proposed a conceptual framework, SMACTR, for developing an internal audit for algorithmic accountability throughout the machine learning development cycle\amdelete{from scoping and mapping to artifact collection and testing and lastly, reflection and post-audit}. The proposed methodology is general-purpose and comprehensive, involving other documentation and checklists discussed in this section (like model cards and datasheets), but this general-purpose methodology may be complicated, expensive and time-consuming to implement, perhaps prohibitively so for teams with limited bandwidth such as the analytics division of a public sector organization. Of note, the SMACTR methodology does not focus on issues of validity. For a given class of problems (e.g., predictive analytics for decision support) there are a set of common validity issues and questions that can be detailed and re-used across contexts. Doing so would complement the SMACTR methodology.

 Based on impact assessments in other domains like construction, algorithmic impact assessments (AIAs) require algorithm developers to evaluate the impacts of the proposed algorithm on society at large and particularly on marginalized communities \cite{reisman2018algorithmic, janssen2020approach, metcalf2021algorithmic}. In 2019 Canada made it compulsory for a government agency using an algorithm to conduct an algorithmic impact assessment \cite{canada2019directiveAIA}. A comprehensive AIA will likely need to involve deliberation about validity issues since an invalid algorithm may very well cause adverse impacts. Related to AIA is the UK Government's Data Ethics Framework which asks practitioners to perform a self-assessment of their transparency, fairness, and accountability \citep{uk_data_ethics_framework}.  \amdelete{The framework additionally asks questions about five ``specific actions": public benefit and user need; diverse expertise; legal compliance; data limitations; model limitations; and policy implications. Regarding public benefit and user need, these questions ask the respondent to consider the benefits of the project and the unintended or negative consequences. The framework asks respondents to articulate user needs in a three bullet point structure that specifies the role (e.g., ``UK resident" or ``data analyst in emergency response"), what the need is, and why they need to do this. Regarding data limitations, the framework asks whether historical bias, selection bias, or proxies for sensitive/protected attributes may be present in the data and what bias mitigation measures have been taken.}
 \amedit{The framework asks the respondent to identify user needs, consider both the benefits and unintended/negative  consequences of the project, and to assess whether historical bias or selection bias may be present in the data.} This framework is helpful in its breadth and specificity. However, the framework does not address core validity issues like proxy outcomes.

A number of toolkits are available to  visualize the performance metrics and tradeoffs therein of algorithmic models. \amdelete{\citet{yu2020keeping} propose a two-step method to communicate tradeoffs to algorithm designers that involves first generating a family of predictive models and subsequently plotting their performance metrics.} Visualization software has been developed \amedit{to communicate tradeoffs to algorithm designers \citep{yu2020keeping}} and to display intersectional group disparities \cite{cabrera2019fairvis}. A number of fairness/ethics toolkits and code repositories are available to help researchers probe model disparities and explore potential mitigations \cite{adebayo2016fairml, bellamy2018ai, saleiro2018aequitas}.

A strain of the literature develops pedagogical processes for improving educational instruction of ethics issues in data science curriculum.  \citet{shen2021value} proposed a toolkit, Value Cards, to facilitate deliberation among computer science students and practitioners.  \amdelete{Drawing upon the model card and fairness checklist literature \cite{mitchell2019model, madaio2020co}, the Value Cards toolkit includes model cards, persona cards describing stakeholders, and checklist cards.They evaluate the use of Value Cards in a classroom context, posing the question of how this extends to practitioners for future research.} The Value Cards largely focus on tradeoffs between performance metrics, stakeholder perspectives, and algorithmic impacts. \citet{bates2020} describes the experience of integrating ethics and critical data studies into  a masters of data science program.

 Guides for best practices in selecting a predictive algorithm for high-stakes settings have been proposed for public policy and healthcare settings \citep{hbrguide_2017, fazel201810ptguide}. For instance, \citet{hbrguide_2017} discuss conceptual issues such as target specification, measurement issues, omitted payoff bias, and selective labels. Our work connects these issues, among others, to established concepts of validity from the social sciences. 
\amdelete{\citet{amershi2019guidelines} use a four-phase process to devise best practice guidelines for the design of human-AI collaborative technologies. 
Their first phase conducts an ad hoc review of design recommendations and systematically condenses them into an initial list of guidelines using affinity diagrams. The second phase refines this list using a heuristic evaluation internally with the research team that asks the participant to apply the guidelines in an example task to evaluate a design interface. The research team discussed points of confusion, redundancy, or inapplicability (that is, guidelines that did not apply to any of their selected sample tasks) and made updates to the list accordingly. As part of this phase, they also specified criteria to ensure consistent formatting of the guidelines.  The third phase conducts a user study to solicit user feedback on the heuristic evaluation which was used to again refine the list of the guidelines. The final phase consisted of an expert validation of the finalized guidelines.}

\subsection{Reliability}

As mentioned earlier, \textit{``reliability is the extent to which measurements are \emph{repeatable} --- when different persons perform the measurements, on different occasions, under different conditions, with supposedly alternative instruments which measure the same thing''} \citep{drost2011validity}. \amedit{Reliability encompasses reproducibility.}
Reliability is also defined as the \emph{consistency} of measurement \citep{bollen1989structural}, and the \emph{stability} of measurement results over a variety of conditions \citep{nunnally1994psychometric}. \amcomment{Are these alternative in the sense that they offer substantively different definitions? or just alternative language for the same concepts?} Reliability is necessary but not sufficient to ensure validity. That is, reliability of a measure does not imply its validity; however, \khedit{a highly unreliable measure}\khdelete{a measure that is not reliable} cannot be valid \citep{nunnally1994psychometric}.

\citet{drost2011validity} enumerates three main dimensions of reliability: equivalence (of measurements across a variety of tests), stability over time, and internal consistency (consistency over time). 
There are several general classes of reliability considerations:
\begin{itemize}
    \item \textbf{Inter-rater reliability} assesses the degree of agreement between two or more raters in their appraisals. \khedit{Low inter-rater }\khdelete{Inter-rater un}reliability could be a potential concern in human-in-the-loop designs where \khdelete{a} human decision-maker\khedit{s} receive\khdelete{s} the predictions of \khedit{a}\khdelete{the} ML model, and interpret\khdelete{s} \khedit{them}\khdelete{it} to reach \khdelete{a }\khedit{the} final decision\khedit{s}. \khcomment{Can we say more here about the circumstances under which this might be a potential concern? I'm thinking it might be important to clarify this. For instance, low inter-rater reliability can sometimes be a symptom of underlying validity issues in an algorithmic system. For instance, is the underlying construct that the algorithmic system is trying to predict fundamentally contested or extremely, irreducibly uncertain?}
    \item \textbf{Test-retest reliability} assesses the degree to which test scores are consistent from one test administration to the next. Population shifts~\citep{quinonero2008dataset}, feedback loops~\citep{ensign2018runaway}, and strategic responses~\citep{hardt2016strategic} are among the threats to the test-retest reliability of risk assessment instruments.
    \item \textbf{Inter-method reliability} assesses the degree to which test scores are consistent when there is a variation in the methods or instruments used. For example, suppose two different models are independently trained to predict the risk of default by loan applicants. Inter-method reliability assesses whether these models often reach similar predictions for the same loan applicants. Another area in which inter-method reliability is applicable to ML is the extent to which an ML model can reproduce the decisions made by human decision-makers.
    \item \textbf{Internal consistency reliability}, assesses the consistency of results across items within a test. Models that make significantly different predictions for similar inputs  may violate this notion of reliability. 
\end{itemize}

\khedit{Efforts in emerging areas such as MLOps focus on}\khdelete{To our knowledge, there are few tools}\khdelete{centered around} \khedit{ the development of practical tools to assess and ensure} \khdelete{assessing and ensuring} the reliability of data-driven predictive analytics \citep{kreuzberger2022machine,shankar2021towards,zaharia2018accelerating}. \khedit{While these efforts are still in their infancy,}\khdelete{However,} there is a growing body of work pointing to \khedit{an urgent}\khdelete{the} need for \khedit{better tooling}\khdelete{such artifacts} \citep{kreuzberger2022machine, shankar2021towards}. For example, \citeauthor{veale2018fairness} identified key challenges for public sector adoption of algorithmic fairness ideas and methods, highlighting the risks posed by changes in policy, data practices, or organizational structures \citep{veale2018fairness}. Focusing on the private sector, 
\citet{holstein2019improving} identified what large companies need to improve fairness in machine learning, highlighting the need for
 ``domain-specific frameworks that can help them navigate any associated complexities." 
In addition to the above changes, feedback loops and strategic responses can induce population shifts, also known as distribution shift or dataset shift \cite{moreno2012unifying}. The literature on data shift concerns the fast detection and characterization of distribution shifts, including distinguishing harmful shifts from inconsequential ones \citep{rabanser2019failing, ashmore2021assuring}. An active area of research in machine learning aims to design learning algorithms that make accurate predictions even if decision subjects respond strategically to the trained model (see, e.g., ~ \citep{dong2018strategic,hardt2016strategic,  mendler2020stochastic, shavit2020strategic, hu2019disparate}). Generalizing such settings, \citet{perdomo2020performative} propose a framework called \emph{performative predictions}, which broadly studies settings in which the act of predicting influences the prediction target. 

While our work focuses on validity issues, we hope that it serves as a jumping off point for future work on reliability artifacts for predictive analytics.

\section{Threats to validity of predictive models} \label{section: challenges}

This section delves into common challenges that jeopardize validity. We organize these challenges into three groups: population misalignment, attribute misalignment, and target misalignment. We connect these groups to notions of validity from the social sciences mentioned in Section~\ref{section: related}. 

\hhcomment{Question: If we choose the wrong optimization objective function for optimization (e.g., measure of accuracy), is that part of validity issues? Which subsection would that fall under?}
\amcomment{I was thinking about this too. I had dropped objective function from the submission draft for the sake of time because I need to think/read a bit more about how to describe this, but I definitely think we should include it at some point. If you have something in mind now, would be great to add to the submission. 
In terms of where it would fit it, maybe target misalignment but this feels like a bit of stretch. Maybe it should be its own category? Objective misalignment?  Are there other related issues we might group—like functional misspecification?}
\khcomment{Agreed, it would be good to pause and revisit which notions of validity are included/excluded in this section, when there is more time.}


\subsection{Attribute Misalignment}
To make meaningful predictions, we must have data on pertinent predictive factors, ideally ones for which we can point to evidence supporting the claim that they are relevant to the predictive task at hand. The choice of which features to use in prediction has clear implications for internal, external, and construct validity. If there is no plausible causal path between the target and a feature such that any correlation is entirely spurious, the inclusion of the feature immediately challenges internal and external validity. Additionally, it can fail tests of face validity. A particularly pressing example of a prediction task that lacks face validity is the use of images of human faces to purportedly ``predict" criminality \citep{wu2016criminalfaces}, because an extensive body of research has disproved the pseudoscience of physiognomy and phrenology \cite{stark2021physiognomic}. 

\hhedit{Note that validity does not require all predictive factors to have a \emph{direct} causal relationship to the target variable.} \hhdelete{predictive factors need not be causal.} For instance, race is a well-established risk factor for COVID-19 related mortality, although \hhedit{the causal pathways through which race and COVID-19 mortality interact are not well-understood} \cite{tai2020disproportionate, mackey2021racial}. \hhdelete{The relationship may instead be associative if}\hhedit{One plausible pathway is that} race is \hhedit{causally} associated with access to healthcare, and access has a causal effect on health outcomes \cite{gray2020covid, mackey2021racial}. \hhdelete{Regardless of the underlying relationship}\hhedit{Given the existence of such plausible causal connection}, race is often invoked as an important risk factor to weigh in allocation of COVID-19 mitigation resources \citep{schmidt2020lawful, wrigley2021geographically}. 

\subsection{Target Misalignment}
In practice there is often considerable misalignment between what humans intended for the algorithm to predict and what the algorithm actually predicts. These issues of construct invalidity can lead to undesirable results after deploying the predictive algorithm. 

In many settings, the desired prediction target is not easily observed\khedit{, and} so a proxy outcome is used in its place. For the pre-trial release task in the criminal justice setting, the desired prediction target may be criminal activity, but it is not possible to directly observe all criminal activity. Instead, algorithm designers have used proxy outcomes like re-arrests or re-arrests that resulted in convictions \cite{fogliato2021validity, bao2021s}. \hhdelete{Measurement error}\hhedit{The use of proxies} in this setting is particularly problematic because there are documented biases in the criminal justice system, such as racial disparities in who is likely to be arrested \cite{alexander2011new}. These systematic biases mean the predictions are not predicting who may commit a crime but instead are predicting who may be arrested. In healthcare contexts, medical costs are sometimes used to proxy health outcomes. However, due to racial bias in quality of healthcare, these proxies systematically underestimate the severity of outcomes for black patients \cite{obermeyer2019dissecting}. In other settings further complications arise when the objective of the decision making task is a function of multiple desired prediction targets. For instance, in the child welfare screening setting decision makers may want to reduce both the risk of immediate danger and the long-term risk of neglect.  When the algorithm is constructed to only focus on one target, then we may suffer \emph{omitted payoff bias} if the algorithm performs worse in practice on the combined objectives than anticipated from an evaluation on the singular objective  \citep{kleinberg2018human}.



 Often we only observe outcomes under the decision taken--that is, we have bandit feedback \cite{swaminathan2015batch}. Prediction tasks in such settings are counterfactual in nature, in the sense that we would like to predict the outcome under a proposed decision \cite{coston2020counterfactual}. An algorithm trained to predict outcomes that were observed under historical decisions will not provide a reliable estimate of what will happen under the proposed decision if the decision causally affects the outcomes. For instance, in a child welfare screening task the goal is to predict risk of adverse child welfare outcomes if no further action is taken (``screened out" of investigation). Investigation can impact the risk of adverse outcomes if the welfare agency is able to identify family needs and provide appropriate services. A predictive algorithm that is trained on the observed outcomes without properly accounting for the effect of investigation on the outcome will screen out families who are likely to benefit from services \cite{coston2020counterfactual}. When we have measured all factors jointly affecting the decision and the outcome, we can address treatment effects by training a counterfactual prediction model \cite{coston2020counterfactual, Schulam17reliable}. When some confounding factors are unavailable for use at prediction time, as long as we have access to the full set of confounding factors in a batch dataset available for training, then we can properly account for any treatment effects in the bandit feedback setting \cite{Coston2020runtime}. In settings where we have unmeasured factors in both the training and test data, we can predict bounds on the partially identified prediction target using sensitivity models \cite{rambachan2022counterfactual}.

\subsection{Population Misalignment}
Even if we can justify our choice of predictive attributes and target variable, we can still have validity issues if the dataset does not represent the target population due to selection bias or other distribution shifts.
This \emph{population misalignment} poses a threat to a valid evaluation of the predictive algorithm because performance on the dataset may not accurately reflect performance on the target population. Notably, fairness properties such as disparities in performance metrics by demographic group can be markedly different on the target population. For example, \citet{kallus2018residual} demonstrated in the context of the New York City Stop, Question, and Frisk dataset that significant disparities in error rates persist in the target distribution (all NYC residents) even when there are no disparities in error rates on the data sample (stopped residents). In the consumer lending context \citet{coston2021characterizing} found that predictive disparities computed on the population of applicants whose loan was approved notably underestimated disparities on the full set of applicants. Misalignment between the model's performance during development and performance at deployment are clear threats to predictive and external validity.

Population misalignment occurs in practice often when the dataset examples are selectively sampled (i.e., not randomly sampled) from the target population. In a number of high-stakes settings, outcomes are only observed for a selectively biased sample of the population. In consumer lending, we only observe default outcomes for applicants whose loan was approved and funded \cite{coston2021characterizing}. In criminal justice, we only observe re-arrest outcomes for defendants who are released \cite{kleinberg2018human}. In child welfare screening, we only observe removal from home for reports that are screened in to investigation \cite{chouldechova2018case}.  A common but  potentially invalid approach in such settings is to use the selectively labelled data to both train the predictive model and perform the evaluation, implicitly treating this sample as if it were a representative sample of the target when in reality it is not.

A promising strategy to address selection bias leverages unlabeled samples from the target distribution which are often already available or could be available under an improved data collection practice \cite{goel2021importance}. For instance, in consumer lending the features (the application information) are available for all applicants \cite{coston2021characterizing}. If we believe that we have measured all factors affecting both the selection mechanism and our outcome of interest (i.e., no unmeasured confounding\footnote{Also known as covariate shift \citep{moreno2012unifying}}), methods are available to perform a counterfactual evaluation that estimate the performance on the full population (including both labelled and unlabelled cases) by taking advantage of techniques from causal inference \cite{dudik2011doubly, coston2020counterfactual}. In settings where we suspect there are unmeasured confounding factors, we can still evaluate a predictive model against the current policy if we can identify an exogenous factor (i.e., an instrumental variable) that only affects the selection mechanism and not the outcome \cite{lakkaraju2017selective, kleinberg2018human}.

Another common mechanism under which population misalignment arises is distribution shift due to domain transfer. 
For example, when expanding credit access to a new international market, a company may want to transfer a model of loan default built on its customer base in one country to the new country \cite{coston2019fair}. Because population demographics and other factors may differ between the two countries, the performance of the predictive model in the source country may not be a valid evaluation of the performance we would see in the new (target) country. When unlabeled data is available from the target domain, we may wish to reweigh the source data to make it ``resemble" the target data. Under the assumption that there are no unmeasured confounding factors that affect both selection into the source/target domain and the likelihood of the outcome (known as \emph{covariate shift}), we can use the likelihood ratio as weights to estimate the performance on the target population \cite{bickel2009discriminative, moreno2012unifying}. We can also use the weights to reweigh the training data in order to retrain a model. 


\amedit{In practice and even with extreme diligence, it is generally not possible to ensure perfect population, target, and attribute alignment. For instance, nearly all prediction settings will suffer population misalignment due to temporal differences—the training data is observed in the past whereas the prediction task is in the future. A central question concerns the \emph{degree} of this misalignment. As a first step towards characterizing this, we propose a deliberation process to identify and reflect on sources of misalignment in a given setting.} 


\section{Deliberating over the validity of predictive models} \label{section: deliberation}

\amdelete{
A predictive algorithm is particularly susceptible to the misalignment in population, target, and/or \hhedit{attributes} discussed in \khedit{the} previous section when the design and development process takes a data-first approach. When we start the design and development with the available data, we risk embarking on a path that neglects key goals and considerations of the targeted decision-making setting. 
For example, consider an analyst reviewing data on child welfare cases that were screened in for investigation. She wonders how well the observed features can predict outcomes like removal from home. After training a model, if the analyst observes the model achieves good out-of-sample performance, she may pitch this as a tool that could be used to support screening decisions. However, if the call screening workers intend to make decisions based on the \akdelete{likelihood of}\akedit{immediate risks} to \akedit{the} child\akedit{'s safety} if not investigated, then this model may suffer target misalignment because removal is a proxy for the unobservable quantity of ``\akedit{immediate safety risks}''. We posit that it will be more productive to reason about these issues \hhedit{at the very beginning of the design process.}\hhdelete{when the process begins with and centers around the \hhdelete{human}\hhedit{ideal} decision-making task.   
\hhcomment{Not sure what you mean by human decision-making---are you taking that to be an oracle/ideal?} \amcomment{Yes, I am. I used human to try to make the connection to human-centered design more natural, but it seems this instead added confusion. I wonder if "ideal" will also be confusing to readers at this point before getting to the discussion below on the oracle/ideal. Should we just remove the adjective?}

Inspired by the human-centered design \amdelete{principles} \amedit{philosophy that forefronts human needs and goals}, \khcomment{What does this refer to? i.e., which human-centered design principles? It may be good to make this more specific. Right now, I think this feels similar to saying ``inspired by a few ideas from reinforcement learning''. People may want to know which ideas, in particular.} \amedit{I edited this to more clearly express what the inspiration was in my mind which was based on Dan Norman's book on everyday design. However you and others certainly know this space much better than I do, so please revise if you see fit}
}}

We propose a series of questions centered around validity to evaluate the justified use of algorithms in a given decision-making context. We next present the top-level questions, discussing them in the context of the child welfare and criminal justice settings. We note that the questions presented in this section are intended purely to illustrate the skeleton of an artifact that is guided by our systematization of concepts from validity theory. Outside the scope of the current contribution, 
future work designing specific sub-questions must solicit feedback from \hhedit{stakeholders and }practitioners to ensure the questions are accessible, comprehensible, and useful. 

\subsection{The High-level Structure of A Validity-Centered Protocol}
At a high level, our proposed artifact will consist of five parts. 
Part 1 prompts the description of the decision-making task and constructs of interest.
Part 2, 3, and 4 consists of questions assessing construct validity, internal validity, and external validity. Last but not least, part 5 attempts to contextualize validity concerns within the broader set of considerations around the use of algorithms (e.g., efficiency). In what follows, we briefly sketch each section.  For illustrative purposes, we provide hypothetical responses in the child welfare screening setting. 

\textbf{1. Description of the decision-making task.} 
To center the deliberation around validity\hhdelete{the considerations and goals of the human decision-making task}, the first set of questions require the respondent to describe \hhedit{the key constructs of interest, including} the decision making objective(s), \hhdelete{the circumstances in which the decision is made, including who is making the decision, based on what criteria}\hhedit{the criteria across which the decision is made}, and other decision points surrounding this task. For example, in the child welfare screening setting, the answer may be as follows: \emph{The hotline call worker determines whether to screen in a report for investigation based on details in the caller's allegations and administrative records for all individuals associated with the report. The report should be screened in if the call worker suspects the child is in immediate danger or at risk of harm or neglect in the future. Preceding this screening decision was the decision by an individual (e.g., neighbor, mandated reporter, other family member) to report to the child welfare hotline. If a report is screened in for investigation, the next major decision point is whether to offer services to the family. A decision to screen out is successful when the child is not at risk of harm or neglect.}

\textbf{2. Questions assessing construct validity:} At a high level, construct validity requires understanding the constructs involved (e.g., the ideal target label and attributes influencing it) and the particular cause and effect relationships among them. To assess construct validity, our protocol will include questions about the following types of validity:
    \begin{itemize}
        \item \textbf{Content validity} asks whether the operationalization of each construct of interest serve as a good measure of it. One major approach to assessing content validity is to ask the opinion of expert\hhdelete{ judge}s in the relevant fields. 
        \item \textbf{Convergent validity:} To assess convergent validity, one must \hhedit{assess}: \hhedit{Is there a standard/ground-truth measure for the construct of interest? If yes, how does that correlate with the new measure on the target population?}
        \item \textbf{Discriminant validity:} To assess discriminant validity, one must \hhedit{evaluate the following}: \hhedit{Can one think of a concept that is related but theoretically different from \hhedit{the} construct of interest? If yes, can the proposed measure distinguish between that concept and the construct of interest?}
        \item \textbf{Predictive validity:} refers to the ability of a test to measure some event or outcome in the future. Therefore, to assess predictive validity, we need to ask: \hhedit{Is there high correlation between the results of the proposed measurement and a subsequent related behavior of interest?}
    \end{itemize}

One effective way to prompt the respondent to respond to the above questions is to consider what question(s) they would ask an oracle who could answer anything about the future. In our child welfare example, the answer here could be as follows: \emph{We would ask whether the child will suffer harm or neglect in the next year.} Subsequent questions will refer to the outcomes identified in this question block as ``oracle outcomes"--that is, the outcomes/events the respondent would like to ask an oracle to predict.

\hhdelete{\amcomment{Seems related to predictive validity}
\hhedit{This question aims to prompt deliberations around construct validity.}
To bring into focus issues of prediction misalignment, this question block asks} 
We follow the oracle question with questions about available outcomes in the data, how these available outcomes differ from the oracle outcome(s), and whether any of the previously stated goals are not addressed by the available outcome. These questions direct the respondent to consider for which segments of the population will the oracle and available outcomes be most likely to align and for which segments of the population will the available outcome likely diverge from the oracle outcome. A key question is when the available outcomes are observed. The answer to these questions may illuminate whether measurement error, bandit feedback, or other forms of missingness pertain to this outcome. An example answer in the child welfare screening context \hhedit{can be the following:} \emph{Available candidate outcomes in the data include re-referral to the hotline at a later point (e.g., within six months) or removal of the child from home within a timeframe (e.g., two years). Re-referral is a noisy proxy for the oracle outcome of harm/neglect because a re-referral can occur in the absence of any harm/neglect and, on the flip side, a child may be experiencing harm or neglect even when no re-referral is made. We expect on average a child that is re-referred to be more likely to experience harm/neglect than a child whose case is not re-referred. Re-referral is more likely to occur, regardless of underlying true risk of harm/neglect, for families of color and limited socioeconomic means \citep{eubanks2018automating, USReport2017, roberts2019digitizing}. Re-referral (or lack thereof) is observed for all reports, including those that are screened in and those that are screened out. By contrast, removal from home is only observed for reports that are screened in for investigation \citep{coston2020counterfactual}.}

A subset of the construct validity questions will direct the respondent to focus on issues of bandit feedback and treatment effects. These questions ask the respondent to consider how the decision relates to the outcome, including whether the outcome is observed under all decisions and whether the decision affects the outcome (and in what ways). For example, the respondent may describe the relationship between the decision and outcome in the child welfare screening setting as follows: \emph{The decision is whether to screen in or screen out a case for a child maltreatment investigation. The outcome that is observed for all decisions is whether the child was later re-referred to the child welfare hotline. If the case is screened in, there are additional observed outcomes: Whether the allegations are substantiated upon investigation by a caseworker, whether the family is offered support in the form of public services, and whether the child is later placed out-of-home. These outcomes are observed under screen out only when a later report is screened in for investigation. The call screener's screening decision affects the outcome. For example, 
the decision to screen in a case may decrease the likelihood of observing adverse outcomes if the family receives public services that lead to improved parenting practices.}

\textbf{3. Questions assessing internal validity:} At a high level, internal validity is concerned with the existence of defensible \emph{causal} relationship between features and the target label.
To hone in on issues of internal validity, the respondent must identify available data features that one can plausibly claim are risk factors or protective factors for the ideal oracle outcome. The respondent must additionally provide evidence to support the claim that these are valid risk factors or protective factors for the oracle outcome.  
For instance, a respondent in the child welfare screening setting may identify the following as risk factors and protective factors in the data: \emph{The data contains the results of any prior child welfare investigations, and we may suspect that a child in a case that was previously found to have child neglect in the past may be at risk for future neglect. The data also contains information on how often extended members of the family (such as the grandmother) interact with or care for the child, and regular supervision from a stable guardian may mitigate risk of child harm or neglect.}

\textbf{4. Questions assessing external validity:} External validity is concerned with the generalizablity of the model across persons, settings, and times. \amedit{The question block focusing on external validity contains questions that require the respondent to describe the population for which data is available (\emph{training population}), including provenance, the locale and time period for which data was observed, and whether any of the observations were filtered out of the dataset (e.g., due to missing data issues). The questions similarly direct the respondent to describe the population on which the predictive algorithm will be used (\emph{target population}), including 
the anticipated time frame and geographies for which the predictive algorithm will be deployed. The respondent will also be asked to specify in what ways the training population differs from the target population. In our running child welfare example, the answer may be: \emph{The training population is all reports to the state's child welfare hotline from 2015-2020 that were recorded in the state records system. No reports were knowingly filtered out of the dataset. The target population is all reports to the state's hotline at least for the next five years. The target population likely differs from the training population because of a change in mandatory reporting in mid 2019 that expanded the definition of mandated reporter to include teachers and sports coaches. As a result, the volume of calls to the hotline increased after the policy change and likely includes some reports that would not have been made absent the policy change.}}

\textbf{6. Tradeoffs between validity and competing considerations:}
To prompt deliberation on how to weigh misalignments threatening validity against other considerations (such as efficiency or standardization), the next set of questions requires the respondent to articulate why a predictive algorithm may support decision making and to describe how they anticipate the predictive algorithm to complement the existing tools and information available. To ground this reflection in specifics, this section will ask respondents 
to precisely identify the expected benefits of the algorithm (e.g., improvements in efficiency or uncovering new patterns of risk). Continuing the child welfare example, the answer may be: \emph{We intend for the predictive algorithm to summarize the information in the administrative records which the call screeners typically do not have sufficient time to fully parse. If the administrative records contain additional patterns of risk not captured in the allegations reported by the caller, then we anticipate the predictive algorithm may be able to flag reports that should be screened in but would otherwise be screened out}. 

\textbf{Target respondent:} The respondent(s) we expect to deliberate and document answers to these questions are the individual(s) involved in the process of bringing data-driven algorithms into the decision-making process. These may include (but are not limited to) algorithm developers, data scientists and analysts, those responsible for algorithm procurement, management, frontline decision makers, and community members.

\amdelete{\textbf{Consider the limitations of the predictive algorithm.}} 

\subsection{Protocol as a Mechanism for Transparency, Oversight, Conversation, \& Translation}
We next discuss how we envision a protocol reflecting the above structure, potentially in combination with questions from other existing protocols (e.g., focused around value alignment), can serve as a mechanism for transparency, oversight,  conversation, and translation. 
\begin{enumerate}
    \item \textbf{Protocol as a mechanism of transparency.} 
    A growing body of literature discusses the need to find better ways to empower impacted community members to shape algorithm design \citep{krafft2021action, zhu2018value,martin2020participatory}. However, community members struggle to do this without sufficient insight into the internal deliberation processes. The protocol can help lower these barriers. For example, without the protocol, community members may be limited to assessing the face validity of models. Publicly shared responses to protocol questions may extend community members' knowledge to encompass a wider range of model validity measures that would otherwise be inaccessible or unknown to them. 
    \item \textbf{Protocol as a mechanism for oversight.} If the protocol is reviewed by an independent review board, deliberations around model validity in decision-making could be guided by standards that may reflect and align expectations across practitioners, policymakers, and community members. We draw an analogy to the research Institutional Review Board (IRB), which has a goal of ``protecting [the rights and welfare of] research subjects'' \citep{belmont1978}. An independent review board for this protocol may serve to protect impacted community members, as opposed to `research subjects.'
    However, the review process may be limited by human biases that challenge the consistency or the quality of review across different applications depending on the reviewer's unique set of biases. 

    \item \textbf{Protocol as a mechanism for conversation between multiple stakeholders.} If a diverse set of stakeholders are involved in deliberating and discussing the protocol questions, the protocol could help these conversations reach those who may not typically be involved in making model-level design decisions. For example, in some public sector agencies that use algorithmic decision support tools, frontline decision-makers, organizational leaders, and model analysts may develop beliefs and goals around the use of decision-making algorithms in silo \citep{kawakami2022improving,saxena2021framework}. The process of responding to the protocol questions can introduce opportunities for structured, proactive modes of interactions across workers who might otherwise typically work in isolation. 
    Engaging diverse perspectives in collaborative discussions surrounding the protocol may open opportunities for better understanding and mitigating inter-organizational value misalignments \citep{holten2020shifting} that would otherwise get embedded and reinforced through the model itself. 

    \item \textbf{Protocol as a mechanism of translation to bridge academic-practitioner divide.} Recent research suggests that many of the concepts under the purview of our envisaged protocol may be less deliberately scrutinized by practitioners developing algorithms for decision-making in the real-world \citep{passi2019problem, veale2018fairness}. The protocol may help bridge this divide between the research community and real-world practitioners. For example, this protocol could be a means for the research community to operationalize concerns related to model validity into practical questions that could guide internal deliberation processes in organizations considering the design or use of algorithms for decision-making.
\end{enumerate}


\subsection{Limitations}
Our paper presents an initial step towards translating theoretical validity concepts into considerations for evaluating the justified use of predictive algorithms in practice. We sketch a structure for a deliberation protocol, targeted to guide multi-stakeholder conversations regarding whether or not to develop and use a predictive algorithm. 
Moving forward, we plan to empirically study practitioners' current practices around validity-related concerns. This research effort will help to ground the protocol, for example, by identifying question categories that may benefit the most from further scaffolding. 
Future work should also explore whether subcategories of real-world domains or types of predictive algorithms require additional or alternative considerations around validity.

Importantly, we emphasize that a validity-focused deliberation protocol is \emph{not} sufficient on its own to justify the use of a predictive algorithm. Rather, we see the primary value of such a protocol as a means to structure and scaffold critical conversations among relevant decision-makers. Moreover, validity is just one component of evaluating the justified use of algorithms, alongside considerations related to reliability, value alignment, and beyond. Last but not least, organizations deploying algorithms should iteratively and constantly re-evaluate whether a predictive algorithm’s use is justified, as the conditions for a given algorithm's justification may evolve with time.

The work in this paper was shaped by the authors' perspectives as machine learning, human-computer interaction, and quantitative social science researchers. Additionally, our experiences working with county and state public agencies over several years informed the work. In future work, we will incorporate perspectives from groups not represented among the authors, including impacted community members.

\section{Concluding Remarks} \label{conclusion}

This paper provides a validity perspective on evaluating the justified use of data-driven decision-making algorithms. This perspective unites concepts of validity from the social sciences with data and problem formulation issues commonly encountered in machine learning and clarifies how these concepts apply to algorithmic decision making contexts. We situate the role of validity within the broader discussion of responsible use of machine learning in societally consequential domains.
We illustrate how this perspective can inform and enhance future research by sketching a validity-centered artifact to promote and document deliberation on justified use.

%



\section*{Acknowledgment}


We thank Alexandra Chouldechova and Motahhare Eslami for their insightful feedback on the project.  Any opinions,
findings, and conclusions or recommendations expressed in this material are solely those of the authors.

\clearpage
 \bibliographystyle{IEEEtranN}
\bibliography{main.bib}

\end{document}